\documentclass{svproc}
\usepackage{url}

\usepackage{amsmath}
\usepackage{amsfonts}
\usepackage{graphicx}
\usepackage[]{subfigure}
\usepackage{setspace}
\usepackage{booktabs} 
\usepackage{multirow} 
\usepackage[T1]{fontenc} 
\usepackage{colortbl} 
\usepackage{xcolor} 

\begin{document}
\mainmatter              
\title{Admittance Control-based Floating \\Base Reaction Mitigation for \\Limbed Climbing Robots}
\titlerunning{Admittance Control-based Floating Base Reaction Mitigation}  
%
\author{Masazumi Imai \and Kentaro Uno \and Kazuya Yoshida}
\authorrunning{Masazumi Imai et al.} 
%
\tocauthor{Masazumi Imai, Kentaro Uno, and Kazuya Yoshida}
\institute{Space Robotics Lab (SRL), Department of Aerospace Engineering, \\
Graduate School of Engineering, Tohoku University, Sendai, Japan,\\
\email{imai.masazumi.p2@dc.tohoku.ac.jp\\
\{unoken, yoshida.astro\}@tohoku.ac.jp}}

\maketitle              

\begin{abstract}
Reaction force-aware control is essential for legged climbing robots to ensure a safer and more stable operation. This becomes particularly crucial when navigating steep terrain or operating in microgravity environments, where excessive reaction forces may result in the loss of foot contact with the ground, leading to potential falls or floating over in microgravity. Furthermore, such robots are often tasked with manipulation activities, exposing them to external forces in addition to those generated during locomotion. To effectively handle such disturbances while maintaining precise motion trajectory tracking, we propose a novel control scheme based on position-based impedance control, also known as admittance control. We validated this control method through simulation-based case studies by intentionally introducing continuous and impact interference forces to simulate scenarios such as object manipulation or obstacle collisions. The results demonstrated a significant reduction in both the reaction force and joint torque when employing the proposed method.
\end{abstract}
\section{Introduction}
\begin{figure}[t]
    \centering
    \includegraphics[width=\linewidth,clip]{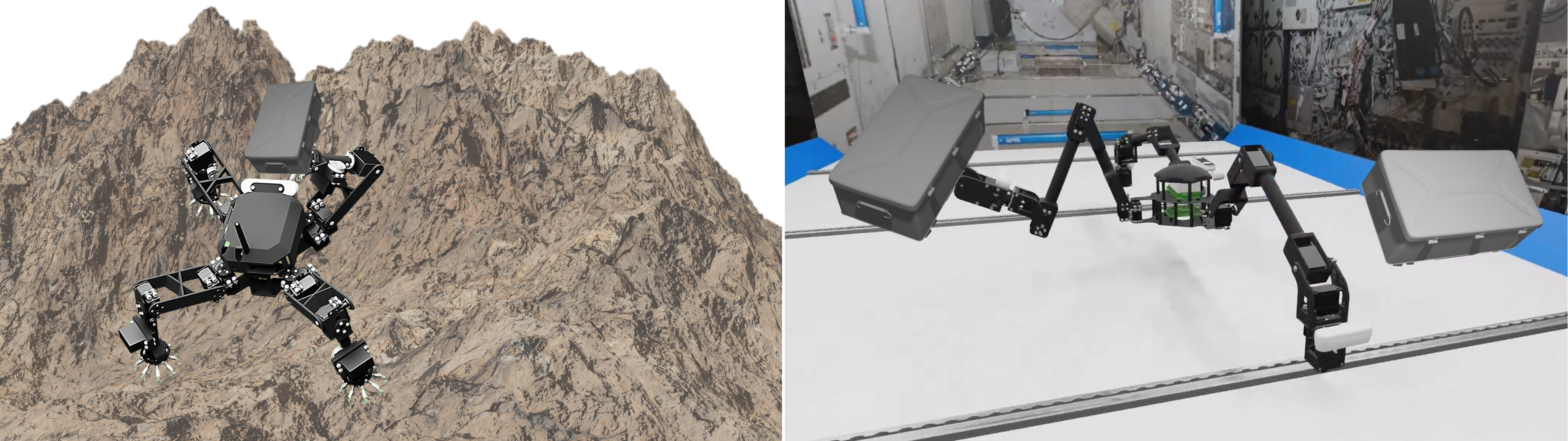}
    \caption{Limbed climbing robots are helpful in an exploration in steep terrain (left) and autonomous tasks in orbital stations (right). In real missions, the robots are supposed to negotiate with the disturbances when manipulating, carrying, or colliding with other objects.}
    \label{fig:fig1}
\end{figure}
Walking robots are gaining attention owing to their potential to operate in harsh environments and replace repetitive tasks. Their utility extends beyond terrestrial applications to include lunar and planetary exploration and operations on orbital stations, thereby expanding their operational range across various gravitational environments~\cite{farley2020mars,mitani2019int}. These robots are equipped with diverse locomotion mechanisms that are tailored to the objectives of specific tasks in various environments.

Among these, limbed-climbing robots, which traverse terrain by grasping environmental surfaces with equipped grippers on their limbs, are particularly effective for exploring cliffs and caves in nature because of their superior ability to traverse rough and inclined terrains. Several such robots have been demonstrated~\cite{parness2017lemur,uno2021hubrobo,tanaka2022scaler,nadanloris}. 
This type of robot can utilize its limbs not only for locomotion but also as a manipulator by turning the grippers on the tips of the limbs into tools that can grasp objects. Therefore, when operating as an exploration robot, it can perform sample mining, collection, and transportation without needing additional mechanisms. Furthermore, robots employing similar concepts of mobility were developed to perform advanced tasks in on-orbit stations by grasping the pre-existing fixtures such as handrails or sheat tracks and moving stably under microgravity~\cite{didot2008eurobot,sun2018prototype} (see Fig.~\ref{fig:fig1}).

However, one of the challenges with such limbed-climbing robots is control stability, owing to the formation of a closed chain caused by multiple grippers interacting with the environment~\cite{wei2016adaptive}.
This issue becomes particularly critical when the robot faces disturbances such as carrying loads after sample collection in steep terrain, colliding with astronauts, or floating obstacles in the on-orbit station. These disturbances generate internal forces on the robot, making stable walking difficult.
Consequently, control strategies that consider internal forces acting on limbed climbing robots are crucial for stable task execution.

\subsection{Related Works}
To enable legged robots to negotiate reactions and disturbances and sustain their stability, compliant control strategies such as Virtual Model Control~\cite{xie2015intuitive} and reinforcement learning-based methods~\cite{hartmanndeep} have been proposed. However, these control strategies require the robot to perform multiple recovery steps, making them unsuitable for limbed climbing robots, whose graspable footholds may be limited and discrete. In addition, many of these force control approaches are based on torque control, which can result in larger and heavier actuators. Such a massive system is undesirable for robots intended for cliff climbing or space applications.

A motion strategy called Reaction-Aware Motion Planning (RAMP) was proposed to mitigate the moments generated from the swinging leg motion for multi-legged locomotion in microgravity~\cite {ribeiro2023ramp}. However, this strategy is not suitable when external disturbances are applied to the robot.

Control strategies employing impedance control have been proposed to reduce the internal forces acting on robots when they form closed chains~\cite{wei2016adaptive}, which method is based on torque control. In another example, {\it admittance control} was proposed to mitigate the effects of buoyancy on underwater-legged robots~\cite{irawan2015center}. Admittance control is a type of indirect force control based on position control~\cite{kosuge1987virtual}, and has also been utilized in research focusing on stabilizing the locomotion of legged robots on uneven terrain~\cite{bjelonic2016proprioceptive}. Nonetheless, these approaches typically apply admittance control only in the vertical direction and lack robustness against forces in the horizontal plane.

\subsection{Contributions}
In this paper, we present a control method based on admittance control to reduce the internal forces for limbed climbing robots, which have a particularly ground-gripping locomotion capability in steep terrain or microgravity environments (see Fig.~\ref{fig:fig1}). In this scenario, the footholds of the robot are usually bonded to the ground, forming a closed chain. Moreover, multi-step reaction mitigation is not preferable. Notably, this scheme requires only force and torque sensor measurements at the foot but no joint-level torque controllability. The main contributions of this study are as follows:
\begin{itemize}
    \item[$\bullet$] A novel control method exploiting admittance control was completely formulated to mitigate internal forces generated when disturbances affect limbed climbing robots.
    \item[$\bullet$] Full dynamic simulation studies showed the proposed method mitigates the reaction forces and balances the joint torques even when external forces are applied to the robot base as well as while locomoting on the steep terrain.
\end{itemize}

\section{Modeling and Method} \label{sec:model_and_control}
%
\subsection{Robot Model}
The equation of motion for a floating-base robot with $N$ limbs is given by (\ref{eq:EOM}):
\begin{equation} \label{eq:EOM}
    \begin{bmatrix}
        \boldsymbol{H}_{\textrm{b}} & \boldsymbol{H}_{\textrm{bm}} \\
        \boldsymbol{H}_{\textrm{bm}}^{\top} & \boldsymbol{H}_{\textrm{m}}
    \end{bmatrix}
    \begin{bmatrix}
        \ddot{\boldsymbol{x}}_{\textrm{b}} \\
        \ddot{\boldsymbol{\phi}}
    \end{bmatrix} + 
    \begin{bmatrix}
        \boldsymbol{c}_{\textrm{b}} \\
        \boldsymbol{c}_{\textrm{m}}
    \end{bmatrix} = 
    \begin{bmatrix}
        \boldsymbol{F}_{\textrm{b}} \\
        \boldsymbol{\tau}
    \end{bmatrix} +
    \begin{bmatrix}
        \boldsymbol{J}_{\textrm{b}}^{\top} \\
        \boldsymbol{J}_{\textrm{m}}^{\top}
    \end{bmatrix} \boldsymbol{F}_{\textrm{e}}
\end{equation}
where, the matrices $\boldsymbol{H}_{\textrm{b}}\in\mathbb{R}^{6\times6}$, $\boldsymbol{H}_{\textrm{m}}\in\mathbb{R}^{n\times n}$, and $\boldsymbol{H}_{\textrm{bm}}\in\mathbb{R}^{6\times n}$, respectively represent the inertia matrix of the robot base, the inertia matrix of the limb, and the coupling inertia matrix between the base and the limb. The terms $\boldsymbol{c}_{\textrm{b}}\in\mathbb{R}^{6}$ and $\boldsymbol{c}_{\textrm{m}}\in\mathbb{R}^{n}$ denote the nonlinear velocity-dependent terms for the base and the limb, while $\boldsymbol{J}_{\textrm{b}}\in\mathbb{R}^{6N\times 6}$ and $\boldsymbol{J}_{\textrm{m}}\in\mathbb{R}^{6N\times n}$ correspond to the Jacobian matrices of the base and the limb, respectively. $\boldsymbol{x}_{\textrm{b}}\in\mathbb{R}^{6}$ and $\boldsymbol{\phi}\in\mathbb{R}^{n}$ denote the positions of the base and joint angles of the limbs, respectively. Furthermore, $\boldsymbol{F}_{\textrm{b}}\in\mathbb{R}^{6}$ and $\boldsymbol{F}_{\textrm{e}}\in\mathbb{R}^{6n}$ indicate external force and moment on the base and the end-effector of the limb, respectively, while $\boldsymbol{\tau}\in\mathbb{R}^{n}$ represent the joint torques exerted by the joint actuators.

\subsection{Admittance Control}
In this section, we introduce the concept of general admittance control, on which the control methodology proposed in this study is based. The admittance control is described as follows:
\begin{equation} \label{eq:admittance}
    \boldsymbol{M}_{\textrm{d}}\Delta\ddot{\boldsymbol{x}}_{\textrm{e,}\textit{i}} + \boldsymbol{D}_{\textrm{d}}\Delta\dot{\boldsymbol{x}}_{\textrm{e,}\textit{i}} + \boldsymbol{K}_{\textrm{d}}\Delta{\boldsymbol{x}}_{\textrm{e,}\textit{i}} = \boldsymbol{F}_{\textrm{e,}\textit{i}}
\end{equation}
where, $\boldsymbol{M}_{\textrm{d}}$, $\boldsymbol{D}_{\textrm{d}}$, and $\boldsymbol{K}_{\textrm{d}}$ indicate the virtual desired inertia, damping, and stiffness matrices, respectively, and by adjusting these parameters, the visual impedance of the end-effector can be controlled. The external forces acting on the $\textit{i}$-th end-effector, $\boldsymbol{F}_{\textrm{e,}\textit{i}}$, were measured using force sensors mounted on each gripper. $\Delta{\boldsymbol{x}}_{\textrm{e,}\textit{i}}$ denotes the deviation between the actual and the equilibrium positions of the $\textit{i}$-th end-effector. By resolving Equation (\ref{eq:admittance}) for the position, the desired position of the $\textit{i}$-th end-effector, ${\boldsymbol{x}}_{\textrm{e,des,}\textit{i}}$ is determined. The desired joint angles were obtained through inverse kinematics based on the desired position relative to the base coordinate system. Subsequently, the robot was controlled by the position control of the joint angles.

Admittance control updates the desired position of the end-effector based on the external forces measured on it. In the context of limbed climbing robots, the end-effectors of the supporting limbs are attached to the environment, meaning that the base position of the robot is updated indirectly in response to the calculated displacement of each end-effector, provided that the grippers remain attached to the ground. Consequently, the indirectly computed pose of the base must maintain kinematic consistency, a requirement that becomes challenging to fulfill when the displacements of multiple end-effectors, each constrained by the environment, require simultaneous updates for each limb. Although admittance control can accommodate this kinematic discrepancy to a certain degree, the resulting internal forces, generated in response to the external forces at the end-effectors, may induce damage to the robot, which forms a closed chain with the environment, leading to the detachment of the grippers.

\begin{figure}[t]
    \centering
    \includegraphics[width=\linewidth,clip]{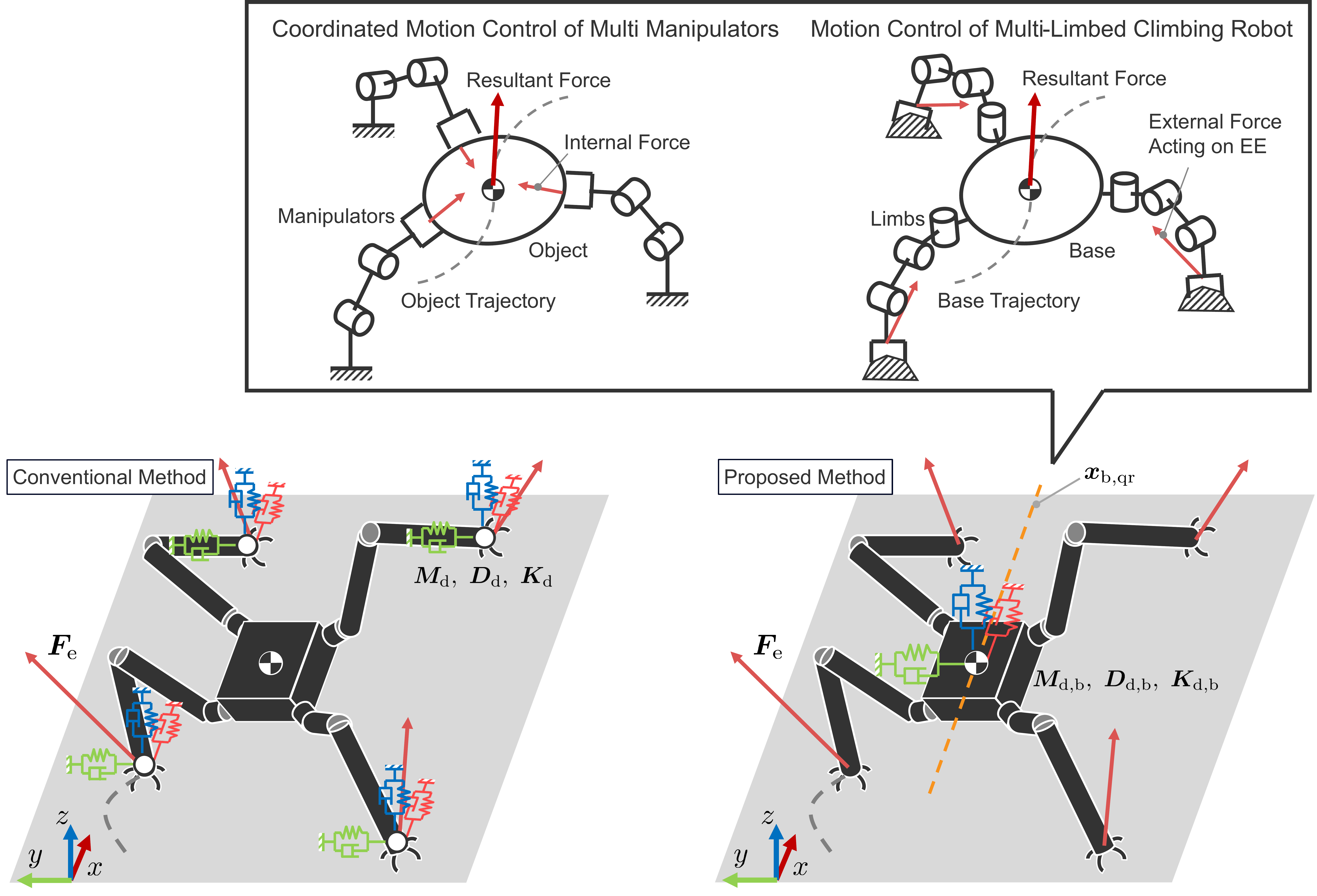}
    \caption{Conceptual diagram of an admittance control for the limbed climbing robot's end-effectors (left) and base COM (right).}
    \label{fig:admittanceForBase}
\end{figure}

Hence, we propose admittance control for the base pose by considering the internal forces exerted on the base. This method is conceptualized based on coordinated motion control principles applied to manipulators. Specifically, we applied force control using a virtual internal model to mitigate the internal forces when multiple fixed-base manipulators manipulated a single object~\cite{caccavale2008six,kosuge2014coordinated}. The scenario in which a floating-base robot walks while grasping the ground surface with its limb end-effectors resembles the manipulation of a single object by manipulators, forming a closed chain between the robot and environment in both cases. 
The internal force acting on the base owing to external forces on the end-effector can be expressed as $\boldsymbol{J}_{\textrm{b}}^{\top}\boldsymbol{F}_{\textrm{e}}$ from the above equation (\ref{eq:EOM}), which implies that our method has the advantage of requiring force and torque sensor measurements only at the end-effector. Thus, the equation representing the proposed admittance control is as follows:
\begin{equation} \label{eq:admittanceForBase}
    \boldsymbol{M}_{\textrm{d,b}}\Delta\ddot{\boldsymbol{x}}_{\textrm{b}} + \boldsymbol{D}_{\textrm{d,b}}\Delta\dot{\boldsymbol{x}}_{\textrm{b}} + \boldsymbol{K}_{\textrm{d,b}}\Delta{\boldsymbol{x}}_{\textrm{b}} = \boldsymbol{J}_{\textrm{b}}^{\top}\boldsymbol{F}_{\textrm{e}}
\end{equation}
where, $\boldsymbol{M}_{\textrm{d,b}}\in\mathbb{R}^{6\times6}$, $\boldsymbol{D}_{\textrm{d,b}}\in\mathbb{R}^{6\times6}$, and $\boldsymbol{K}_{\textrm{d,b}}\in\mathbb{R}^{6\times6}$ indicate the virtual desired inertia, damping, and stiffness matrices, respectively. If $\boldsymbol{x}_{\textrm{b}}\in\mathbb{R}^{6}$ and $\boldsymbol{x}_{\textrm{b,eq}}\in\mathbb{R}^{6}$ denote the actual pose and the equilibrium pose of the robot base, respectively, then $\Delta{\boldsymbol{x}}_{\textrm{b}}\in\mathbb{R}^{6}$ is defined as following (\ref{eq:diffBasePose}).
\begin{equation} \label{eq:diffBasePose}
    \Delta{\boldsymbol{x}}_{\textrm{b}} = \boldsymbol{x}_{\textrm{b}} - \boldsymbol{x}_{\textrm{b,eq}}
\end{equation}
The desired base pose $\boldsymbol{x}_{\textrm{b,des}}$ is computed by solving Eq. (\ref{eq:admittanceForBase}) with respect to the pose. Subsequently, employing the procedure described earlier, the desired joint angles were acquired through inverse kinematics, enabling the robot to be controlled via position control. The application of admittance control for the end-effector and base center of mass (COM) in limbed climbing robots is shown in Fig.~\ref{fig:admittanceForBase}.

\section{Simulation} \label{sec:simulation}
Dynamic simulations were conducted to validate the proposed control method for limbed-climbing robots. ClimbLab, which is an open-source simulator developed by our research group, was used in this study~\cite{uno2022climblab}. ClimbLab facilitates the rapid design and analysis of new planning and control algorithms for limbed climbing robots using MATLAB.
\subsection{Condition}
We employed the HubRobo~\cite{uno2021hubrobo}, a quadruped climbing robot, as a model for our study. The maximum tolerable grasping force was set to 15~N, and gripper detachment from the ground surface occurred when the reaction force acting on the end-effector of the robot in the pulling direction exceeded this threshold.
The contact force between the end-effector and the environment is calculated using a spring-damper model based on the virtual penetration amount $\boldsymbol{\delta}$ and is expressed as
\begin{equation} \label{eq:contactModel}
    \boldsymbol{F}_{\textrm{e,}\textit{i}} = \boldsymbol{K}_{\textrm{s}}\boldsymbol{\delta}_{\textit{i}} + \boldsymbol{D}_{\textrm{s}}\dot{\boldsymbol{\delta}}_{\textit{i}}
\end{equation}
where $\boldsymbol{K}_{\textrm{s}}\in\mathbb{R}^{3}$ and $\boldsymbol{D}_{\textrm{s}}\in\mathbb{R}^{3}$ are the stiffness and damping coefficients, respectively. Assuming $k_{\textrm{s}}$ and $d_{\textrm{s}}$ represent the component in each axis direction of $\boldsymbol{K}_{\textrm{s}}$ and $\boldsymbol{D}_{\textrm{s}}$, the values of these coefficients for the simulation are set as $k_{\textrm{s}} = 10000$ and $d_{\textrm{s}} = 20$. In this study, we employed a simple periodic crawling gait for robot locomotion planning.

To assess the robustness of the proposed control method concerning reaction forces on the robot base, two types of external forces are applied to the base COM during walking: 1) apply an additional mass onto the base, causing an extra gravitational force continuously, and 2) apply multi-times impact forces at the base, which are associated with the following real-world scenarios: 1) the robot loads cargo (assumed to fetch a sample or a measurement device), and 2) a human crew member or obstacle collides with the robot in a space station.
In Case~1), two simulation patterns were executed, demonstrating Earth's and lunar gravity conditions. An additional mass weighing 1kg, approximately half the weight of HubRobo, was placed on the robot during the two gait cycles (five steps).
In Case~2), a microgravity environment was assumed, set at $1\times10^{-6}$~G, to emulate the conditions in space. External forces were applied to the base three times within the time intervals of $1.0 \leq t \leq 1.5$, $4.5 \leq t \leq 5.0$, $7.0 \leq t \leq 7.5$. The vectors representing the external forces for each time interval are defined as $\boldsymbol{F}_{\textrm{b}} = [-5,~0,~0,~0,~0,~0]^{\top}, [0,~5,~0,~0,~0,~0]^{\top}, [0,~-10,~0,~0,~0,~0]^{\top}$
To assess and compare the performance, the PD control without the proposed method was used as the baseline. PD control is also utilized for the position controller in the proposed method. The control parameters used in the simulations are listed in Table~\ref{tab:control_param}. Here, $m_{\textrm{d,b}}$, $d_{\textrm{d,b}}$, and $k_{\textrm{d,b}}$ represent the elements of the diagonal matrices $\boldsymbol{M}_{\textrm{d,b}}$, $\boldsymbol{D}_{\textrm{d,b}}$, and $\boldsymbol{K}_{\textrm{d,b}}$, respectively, and the value of the virtual inertia is fixed at 1 and the values of the damping and stiffness coefficients are determined empirically.
Furthermore, the proportional gain $k_{\textrm{p}}$ and derivative gain $k_{\textrm{d}}$ for the PD control were uniformly set to 50 and 0.2, respectively, across all the simulation cases.
\begin{table}[b]
    \caption{Control parameters for the simulation}
    \begin{center}
    \begin{tabular}{r@{\quad}c@{\quad}c@{\quad}c}
    \hline\rule{0pt}{12pt}
    Parameter & \multicolumn{3}{c}{Value} \\[2pt]
    \cmidrule(lr){2-4}\rule{0pt}{12pt}
     & Case 1.1) 1~G & Case 1.2) 1/6~G & Case 2) \\[2pt]
    \hline\rule{0pt}{12pt}
    $m_{\textrm{d,b}}$ & 1 & 1 & 1 \\
    $d_{\textrm{d,b}}$ & $1\times10^{4}$ & $5\times10^{3}$ & $5\times10^{3}$ \\
    $k_{\textrm{d,b}}$ & $4\times10^{5}$ & $5\times10^{4}$ & $2\times10^{4}$ \\
    \hline
    \end{tabular}
    \end{center}
    \label{tab:control_param}
\end{table}

%
\subsection{Results and Discussions}
\begin{figure}[ht]
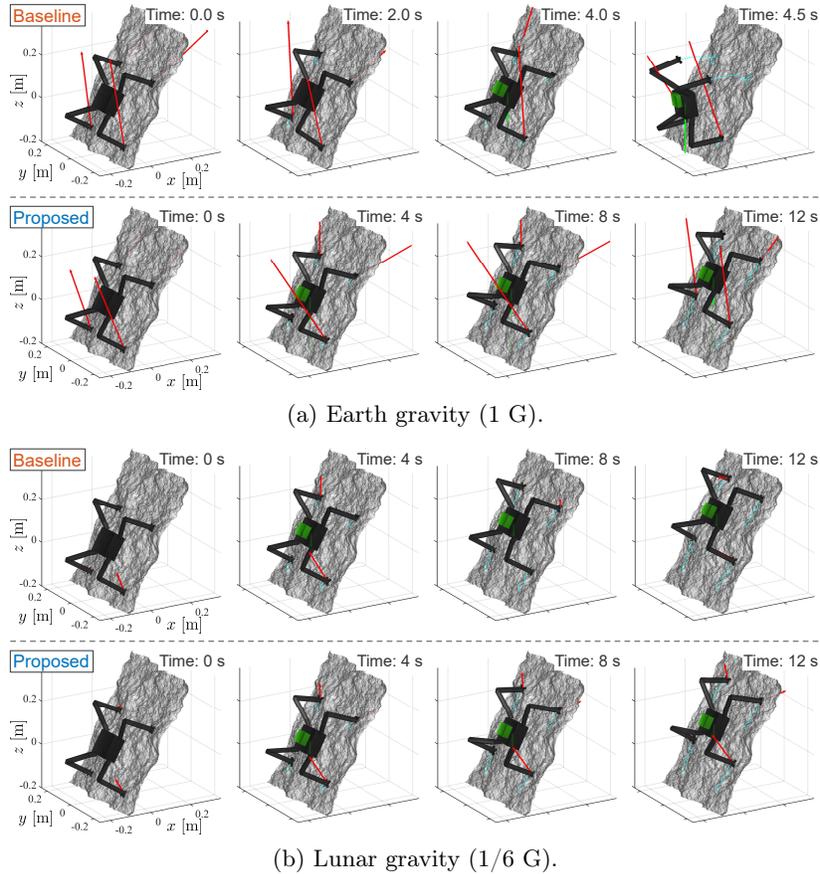

    \centering
    \subfigure{
        \includegraphics[width=0.9\linewidth]{./fig/load_cargo_earthG_sim.pdf}
    }\\
    {\footnotesize (a) Earth gravity (1~G).}\\
    \subfigure{
        \includegraphics[width=0.9\linewidth]{./fig/load_cargo_lunarG_sim.pdf}
    }\\
    {\footnotesize (b) Lunar gravity (1/6~G).}\\
    \caption{Snapshots of loading cargo simulation under (a) Earth and (b) Lunar gravity, comparing the baseline: simple PD control (top) and proposed method (bottom) in each condition. The red and green arrows indicate reaction forces acting on the end-effector and the base, respectively. The light blue curves are trajectories of feet.}
    \label{fig:load_cargo_sim_result}
\end{figure}

The simulation results for cargo loading are presented in Fig.~\ref{fig:load_cargo_sim_result}.
\begin{figure}[t]
    \centering
    \subfigure{
        \includegraphics[width=0.45\columnwidth,clip]{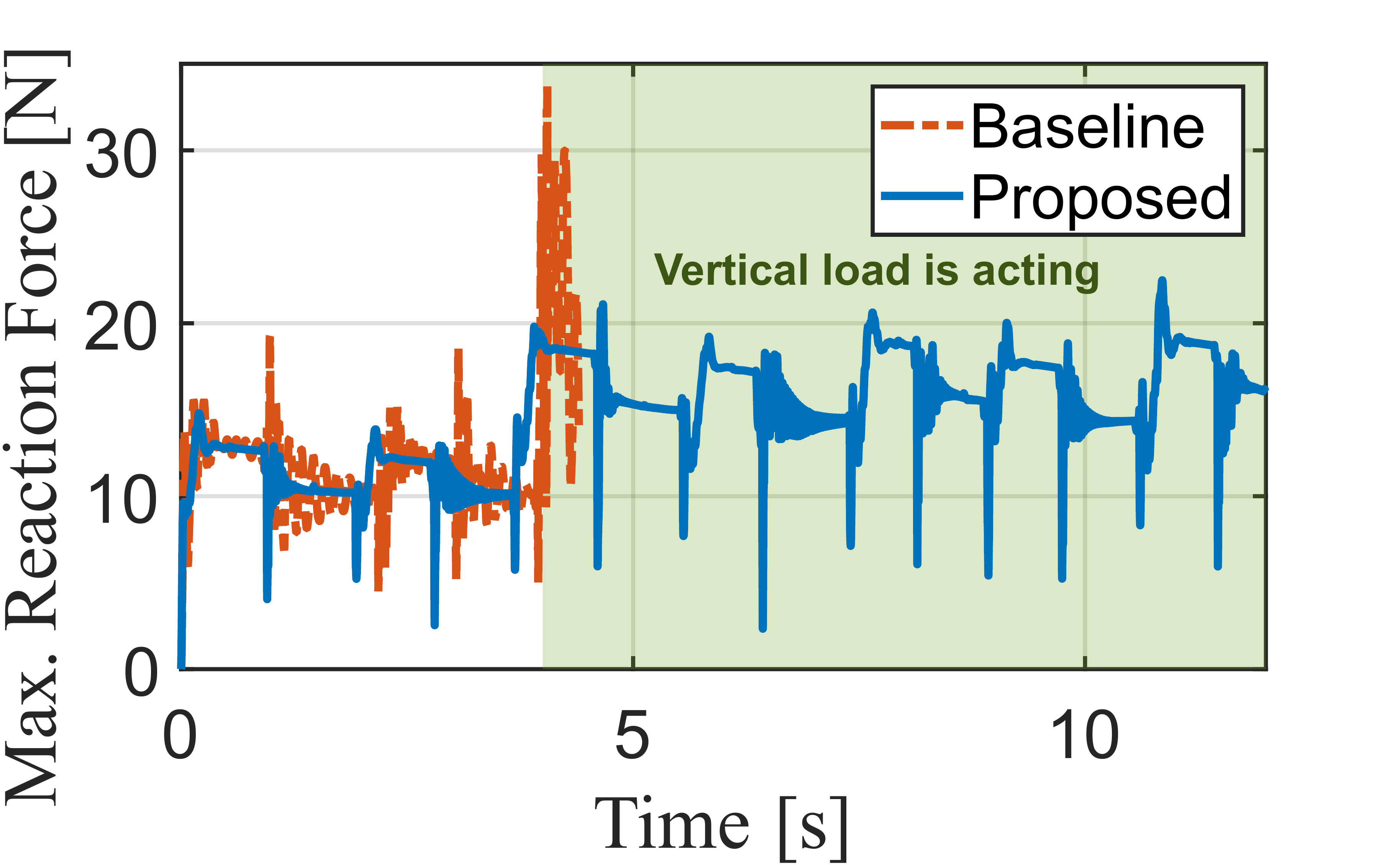}
    }
    \subfigure{
        \includegraphics[width=0.45\columnwidth,clip]{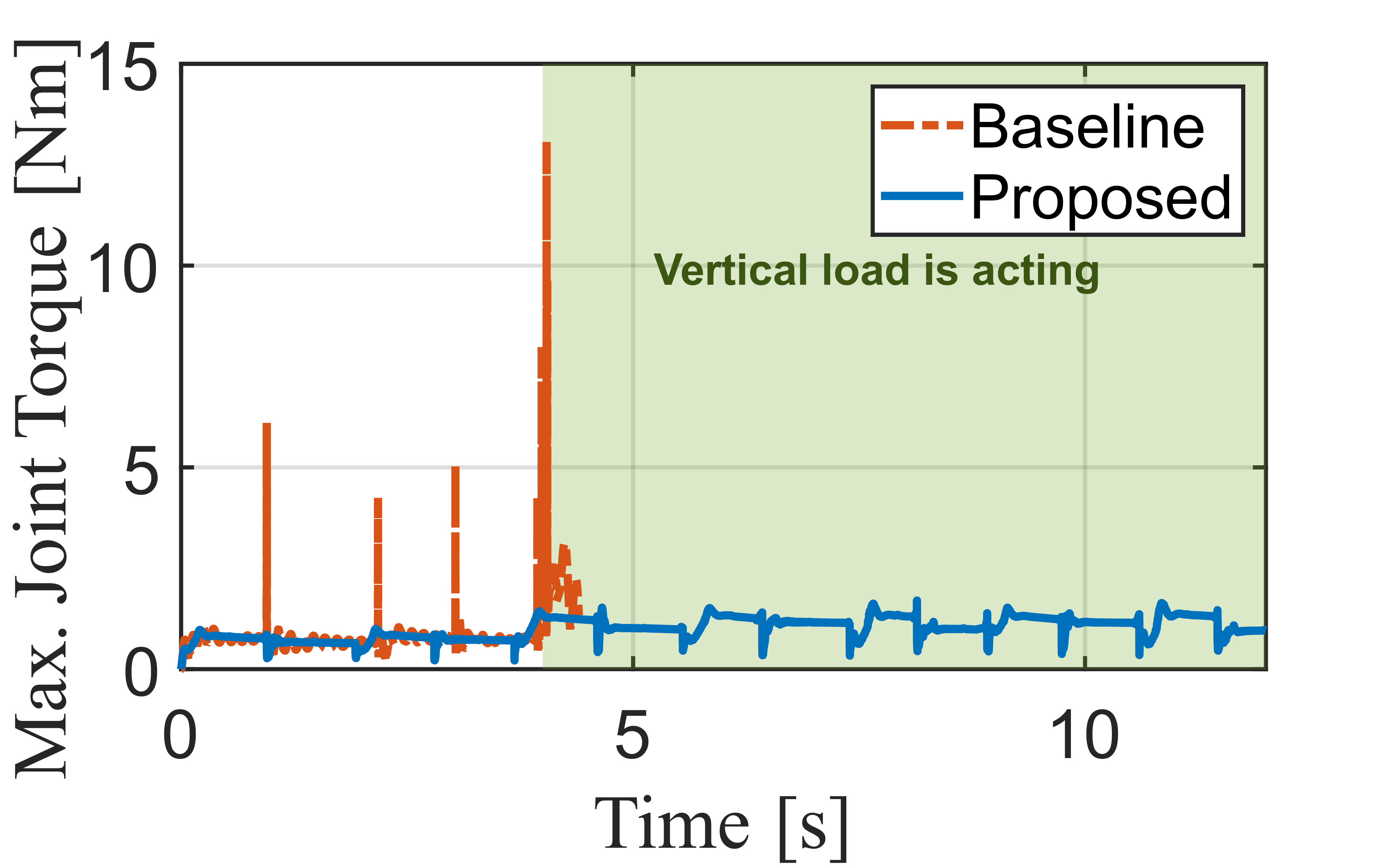}
        \label{fig:load_cargo_earthG_graph}
    }\\
    {\footnotesize (a) Earth gravity (1~G).}\\
    \subfigure{
        \includegraphics[width=0.45\columnwidth,clip]{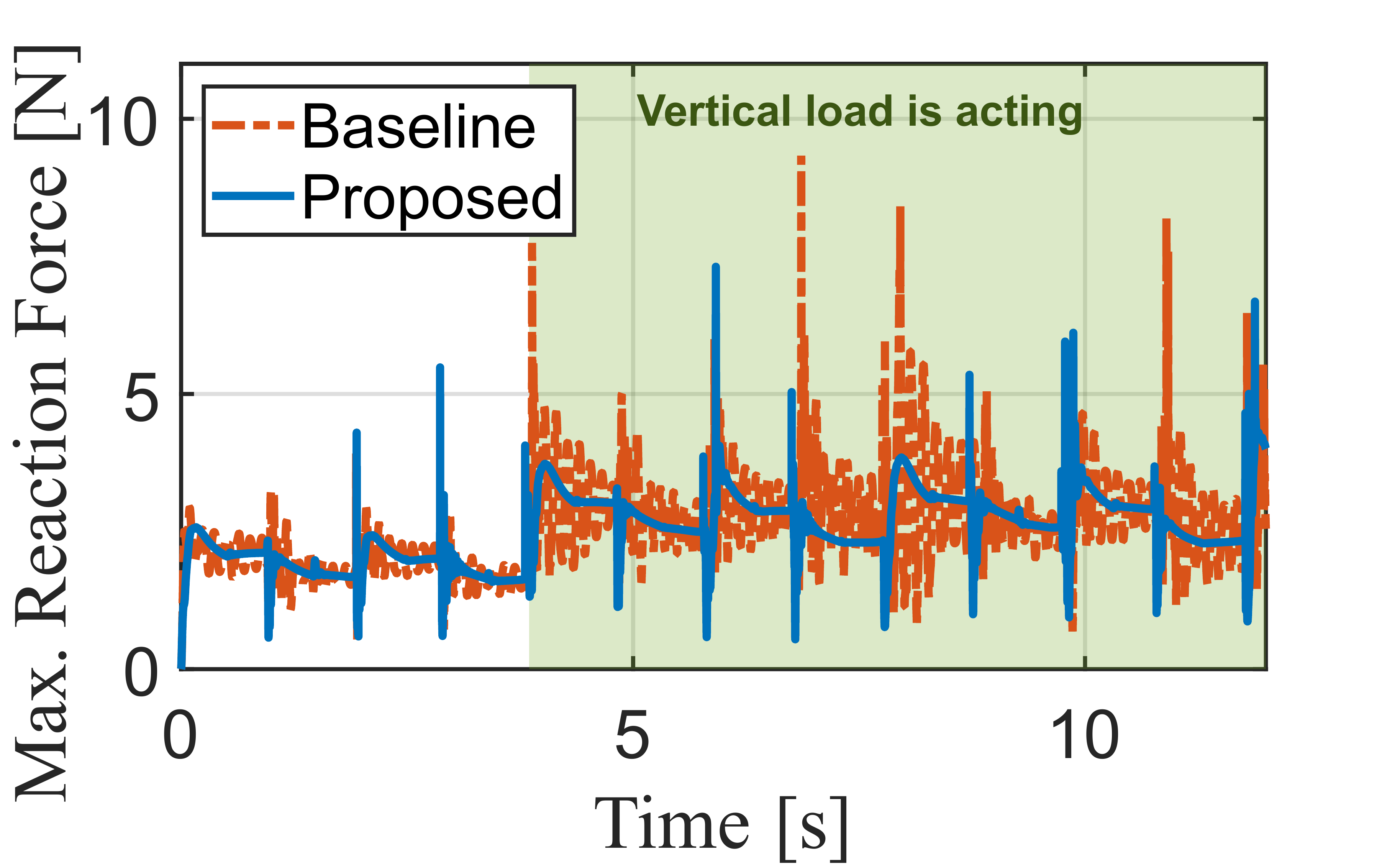}
    }
    \subfigure{
        \includegraphics[width=0.45\columnwidth,clip]{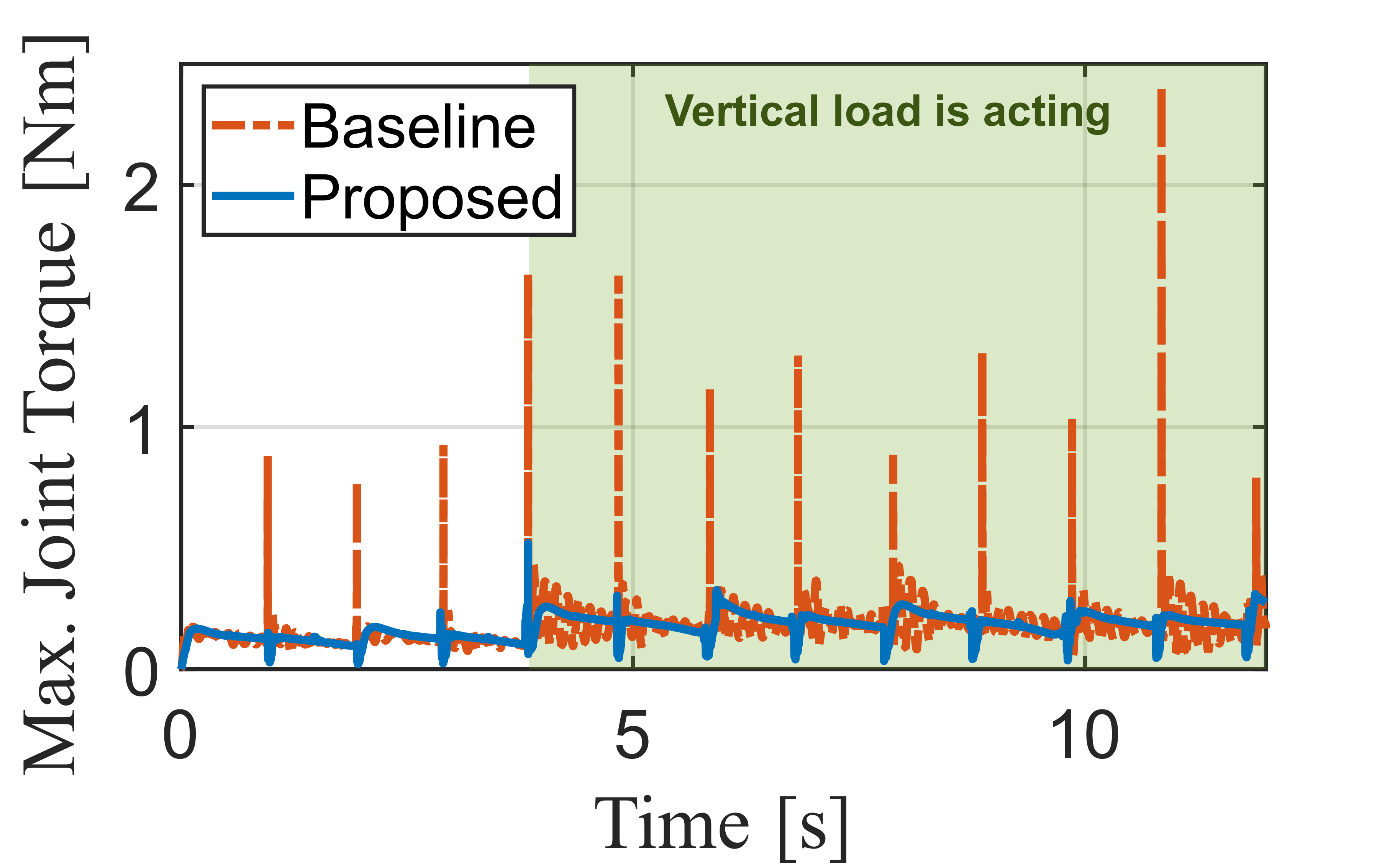}
        \label{fig:load_cargo_lunarG_graph}
    }\\
    {\footnotesize (b) Lunar gravity (1/6~G).}\\
    \caption{Comparison graphs of maximum reaction force (left) and joint torque (right) between the baseline and proposed method under Earth and Lunar gravity.}
    \label{fig:load_cargo_graphs}
\end{figure}
\begin{figure}[t]
    \centering
    \subfigure[\footnotesize Earth gravity (1~G).]{\includegraphics[width=0.45\columnwidth,clip]{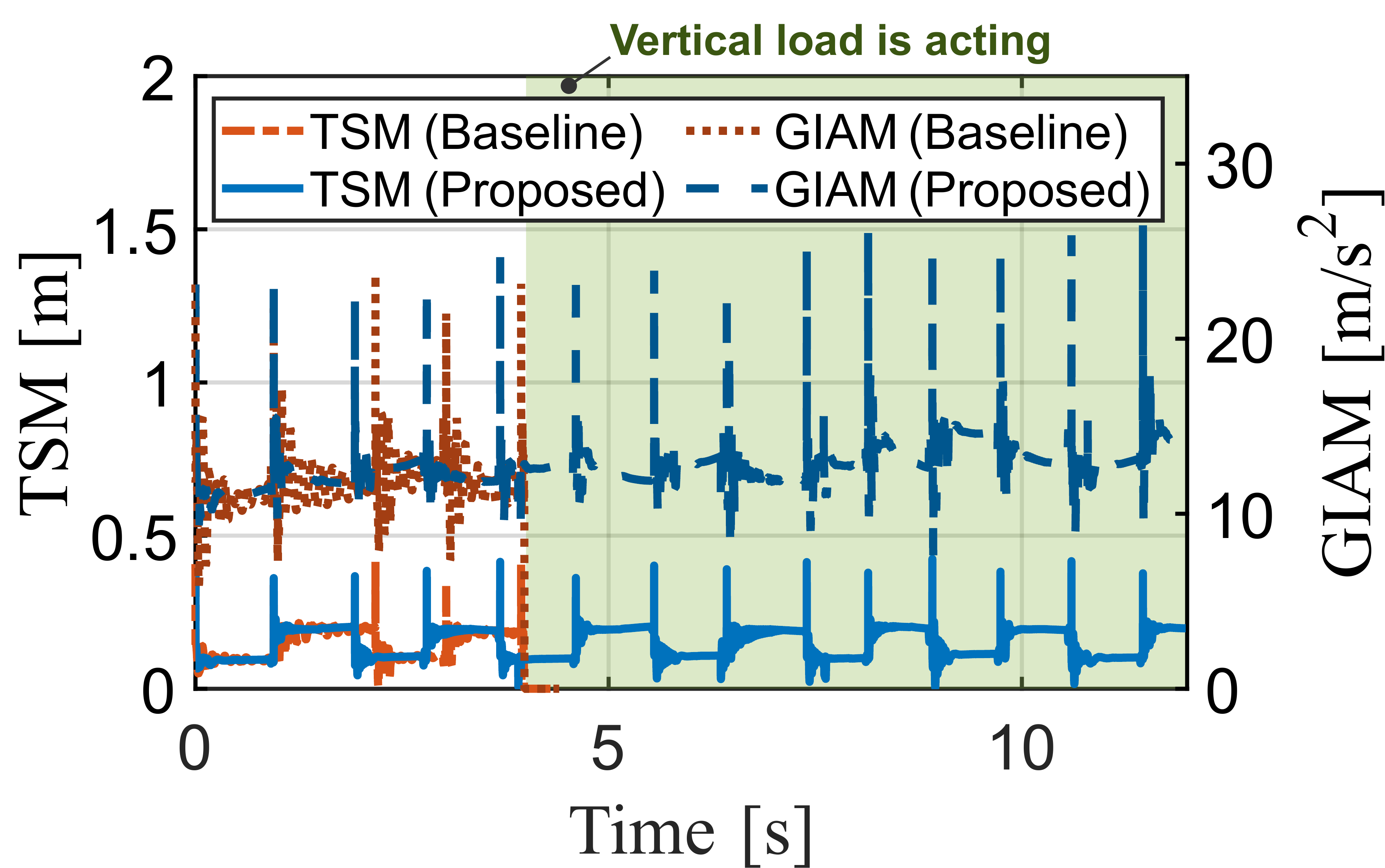}
    \label{left}}
    \subfigure[\footnotesize Lunar gravity (1/6~G).]{\includegraphics[width=0.45\columnwidth,clip]{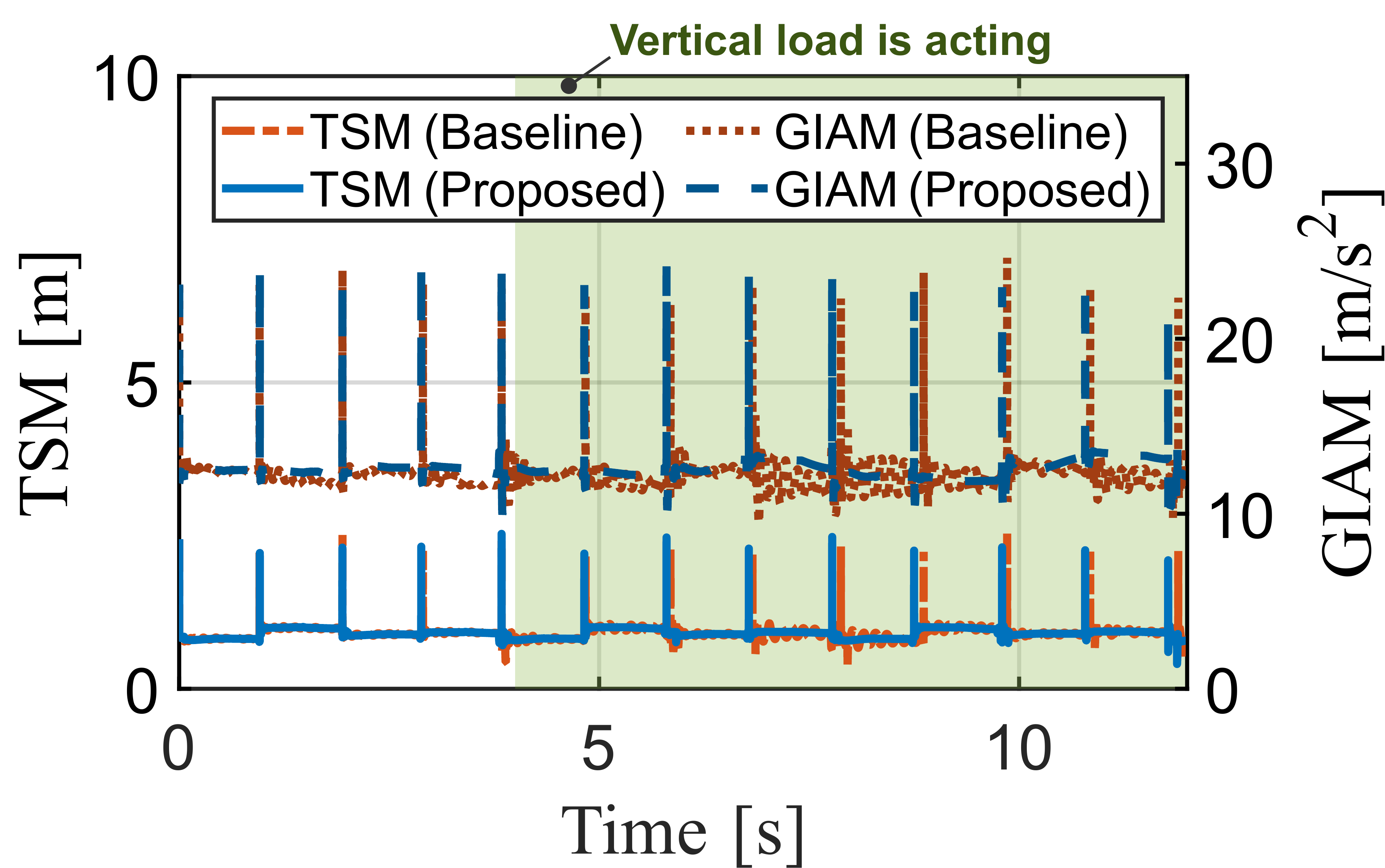}
    \label{right}}
    \caption{Comparison graphs of the TSM (left axis) and GIAM (right axis) between the baseline and proposed method under Earth and Lunar gravity.}
    \label{fig:load_cargo_stability}
\end{figure}
Additionally, Fig.~\ref{fig:load_cargo_graphs} shows a time plot comparing the maximum reaction force applied to the end-effectors and the joint torque of all the limbs between the baseline and proposed control methods.
From Fig.~\ref{fig:load_cargo_graphs}(a), the proposed method effectively reduced the oscillation of the end-effector reaction force under Earth's gravity conditions. This outcome suggests that our proposed control approach offers stable operation compared to the baseline because unstable ground reaction forces can result in gripper detachment for limbed climbing robots. Additionally, as observed in the torque graph, the proposed method eliminated spikes in the joint torque. This indicates that the proposed method balances the joint torque and mitigates the risk of damage caused by the instantaneous torque exerted by the actuator.
Furthermore, we evaluated the stability measures by comparing the simulation results using the Tumble Stability Margin (TSM)~\cite{yoneda1996tumble}, an extension of the stability margin for climbing robots, and the Gravito-Inertial Acceleration Margin (GIAM)~\cite{ribeiro2020dynamic}, a criterion for defining the dynamic equilibrium of climbing robots. Fig.~\ref{fig:load_cargo_stability}(a) shows the time histories of TSM and GIAM in the simulation results with and without the proposed control method. Detachment can also be observed when the TSM and GIAM become zero at baseline, whereas detachment does not occur in the proposed method.

Similarly, the simulation results for cargo loading under lunar gravity are shown in Fig.~\ref{fig:load_cargo_sim_result}(b) and Fig.~\ref{fig:load_cargo_graphs}(b) compares the maximum joint torque and maximum reaction force of all limb end-effectors between the cases with and without the proposed method under lunar gravity.
Fig.~\ref{fig:load_cargo_graphs}(b) shows that the proposed method exhibits more stable reaction forces applied to the end-effectors and joint torque, which is consistent with the trends observed under Earth's gravity conditions, as described above.
From Fig.~\ref{fig:load_cargo_stability}(b), there is no significant difference in TSM and GIAM between the baseline and proposed cases. However, after an external force is applied to the base, fluctuations in TSM and GIAM values are observed in the baseline case, whereas no such fluctuations are observed in the proposed case. These results indicate that the proposed method offers superior stability compared with the baseline method.

In addition to comparing the reaction forces and joint torque, we evaluated the simulation studies from the perspective of energy efficiency using the Cost of Transport (COT).
The COT measures the amount of energy expended per unit movement speed; a smaller COT value indicates more efficient locomotion. The average COT for each case is listed in Table~\ref{tab:COT}.
However, for Case 1) (1G), the average COT value was not available because of the failure to walk using the baseline method. In the other cases, the average COT values for the proposed method were lower than those for the baseline. This suggests that the proposed method effectively balances the joint torques to reduce energy consumption, even when external forces are applied to the robot base.
\begin{table}[b]
    \caption{Control parameters for the simulation}
    \begin{center}
    \begin{tabular}{l@{\quad}c@{\quad}c@{\quad}c@{\quad}c}
    \hline\rule{0pt}{12pt}
    Quantity & Controller & Case 1.1) 1~G & Case 1.2) 1/6~G & Case 2) \\[2pt]
    \hline\rule{0pt}{12pt}
    \multirow{2}{*}{Average COT} & Baseline & - & 1.42 & $8.88\times10^{4}$ \\
      & Proposed & 1.31 & 1.38 & $6.86\times10^{4}$ \\
    \hline
    \end{tabular}
    \end{center}
    \label{tab:COT}
\end{table}

\begin{figure}[t]
    \centering
    \subfigure{
        \includegraphics[width=0.9\linewidth,clip]{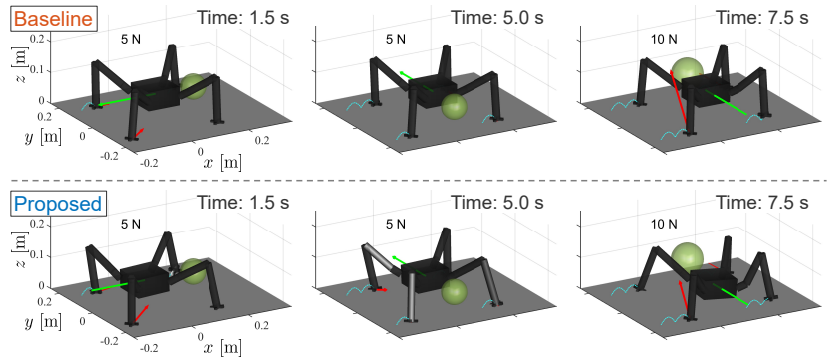}
    }
    \caption{Comparison of simulation results for applying multi-times impulse forces under microgravity (top: baseline, bottom: proposed control).}
    \label{fig:result_impulse_forces_microG}
\end{figure}

\begin{figure}[t]
    \centering
    \subfigure{
        \includegraphics[width=0.45\columnwidth,clip]{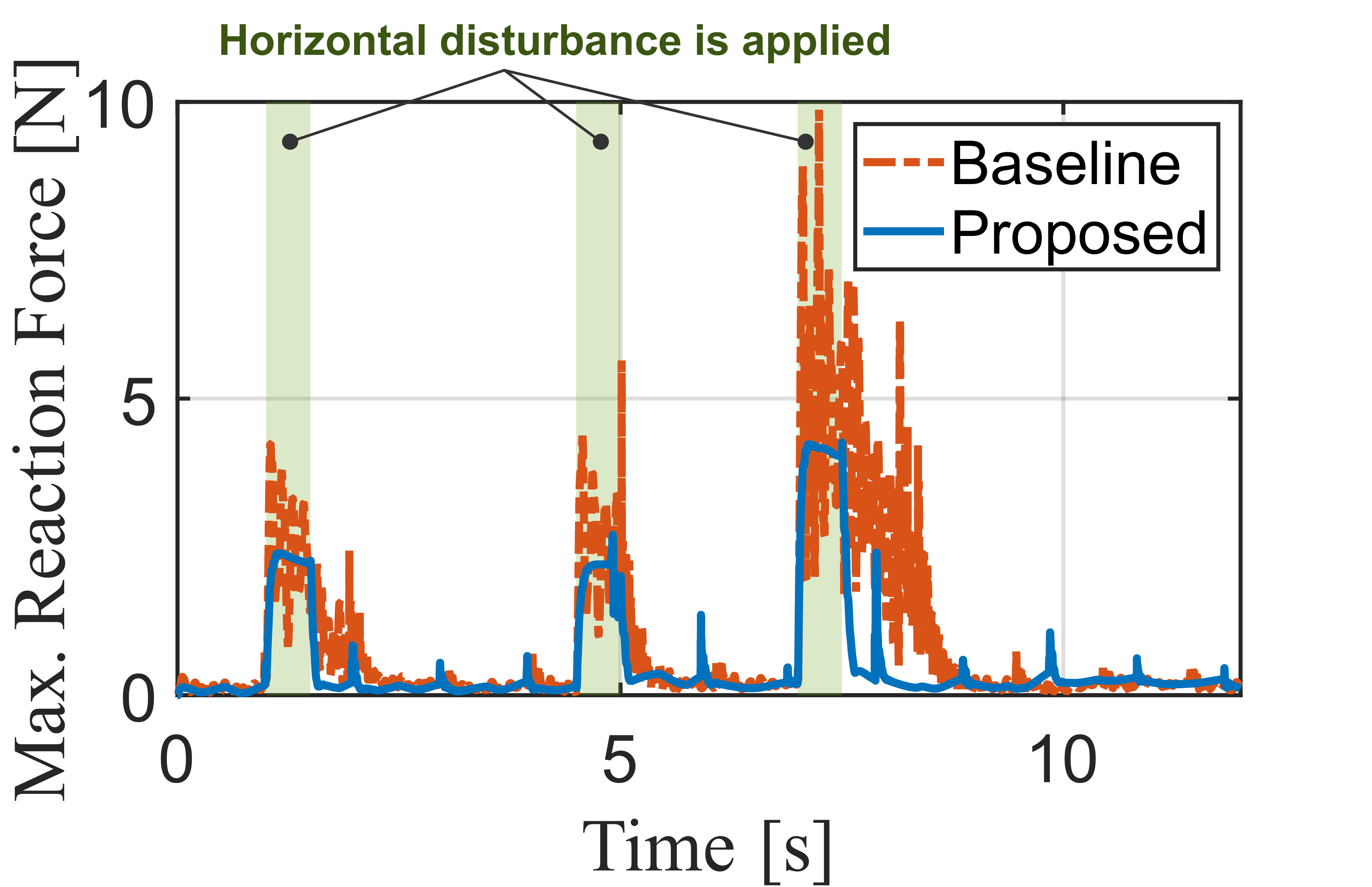}
        \label{fig:impulse_force_microG_max_reaction_force}
    }
    \subfigure{
        \includegraphics[width=0.45\columnwidth,clip]{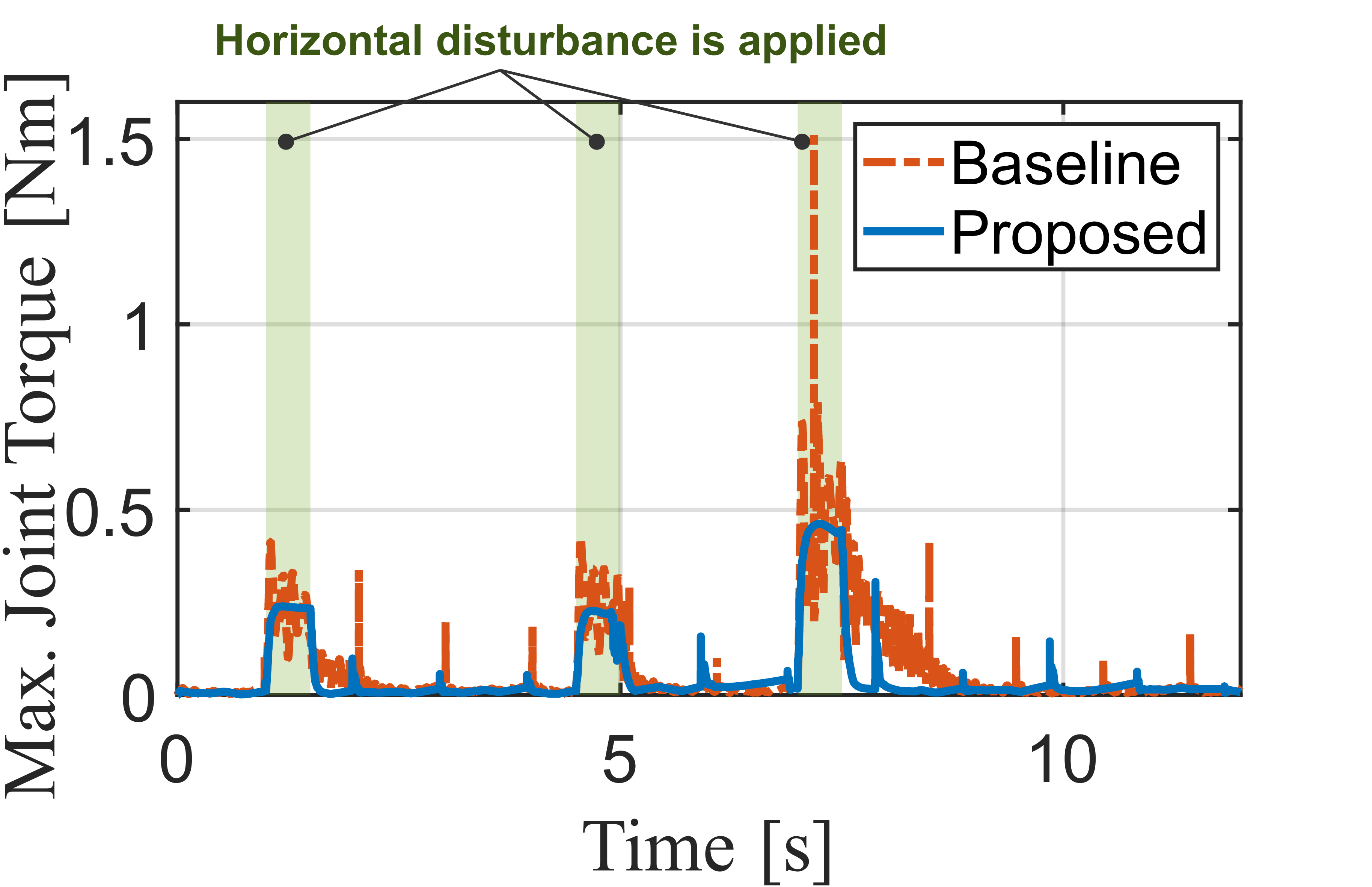}
        \label{fig:impulse_force_microG_max_joint_torque}
    }
    \caption{Comparison graphs of performance between the baseline and proposed method under microgravity.}
    \label{fig:impulse_force_microG_graphs}
\end{figure}

The results of the simulations for Case~2), involving the application of impulse forces to the robot, are shown in Fig.~\ref{fig:result_impulse_forces_microG}. Moreover, Fig.~\ref{fig:impulse_force_microG_graphs} depicts graphs comparing the maximum reaction force of all limb end-effectors and the maximum joint torque between the baseline and proposed control methods in Case~2).
\begin{figure}[t]
    \centering
    \includegraphics[width=.45\linewidth,clip]{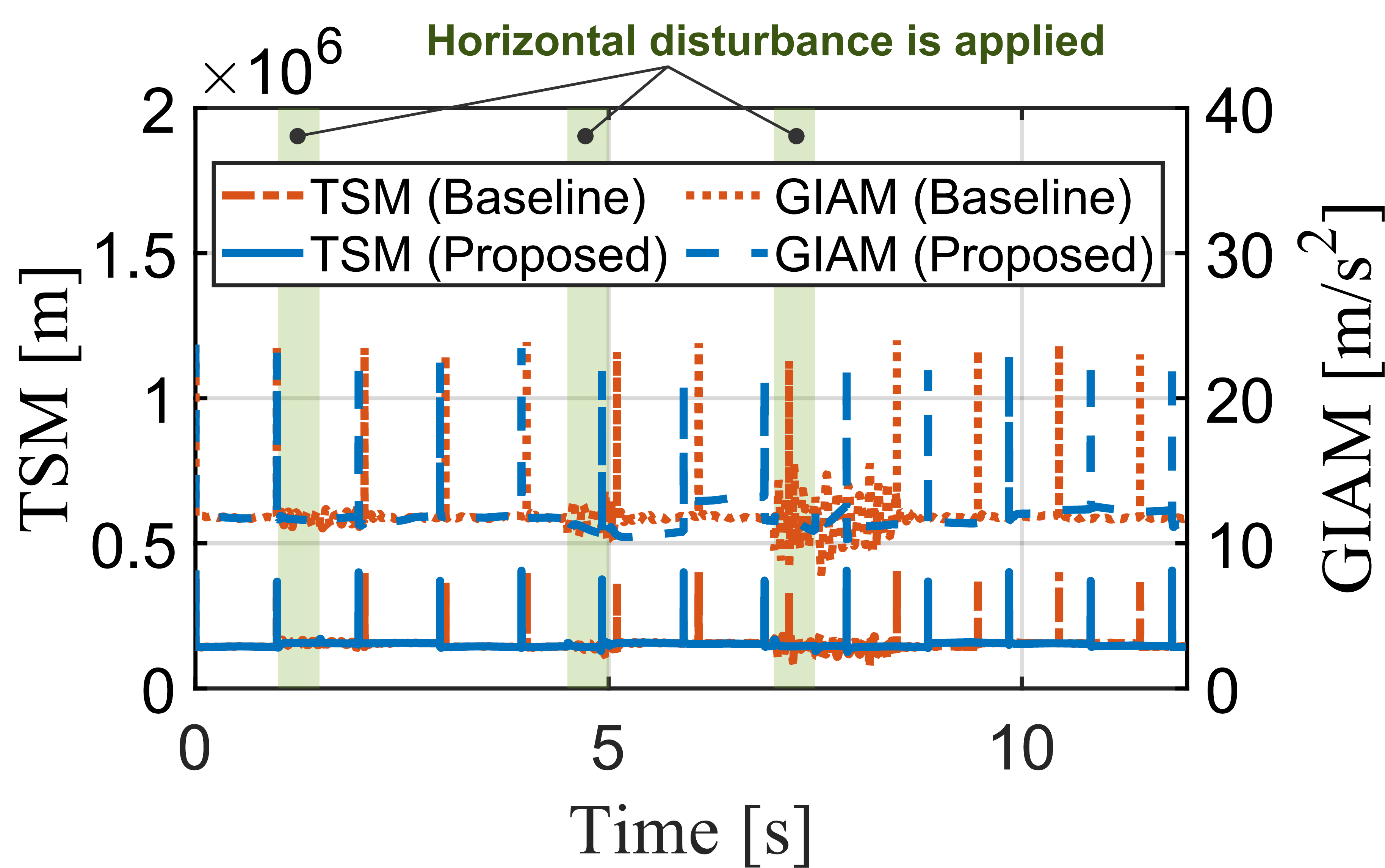}
    \caption{Comparison graphs of the TSM and GIAM between the baseline and proposed method under microgravity.}
    \label{fig:impulse_force_microG_stability}
\end{figure}
In Case~2), we can see that, with the proposed method, the robot moves its base in the same direction as the external forces, whereas in the baseline, the robot maintains its base position even when external forces are applied. Additionally, the proposed method leads to a reduction in both the reaction force on the end-effector and the joint torque exerted by the joint actuator, as shown in Fig.~\ref{fig:impulse_force_microG_graphs}. Consequently, these findings indicate that the proposed method demonstrates high robustness against the reaction forces acting on the robot base compared with the baseline.
Furthermore, as shown in Fig.~\ref{fig:impulse_force_microG_stability}, similar to Case~1), the baseline method exhibited oscillations in TSM and GIAM values after the application of an external force. By contrast, the proposed method did not exhibit such oscillations, demonstrating its superior stability.

\section{Conclusion} \label{sec:conclusion}
In this study, we presented a control strategy for a limbed climbing robot based on admittance control applied to the robot base and evaluated its effectiveness. Our proposed method effectively reduced the ground reaction force on the end-effector when external forces were applied to the robot base, as demonstrated through dynamic simulations. This validation highlights the robustness of the proposed method against external forces acting on the base.

In the future, we will optimize the control parameters to impart impedance characteristics that are conducive to various environments while considering kinematic constraints such as joint mobility limitations.
Additionally, when larger external forces act on the robot, there is a possibility that the joint angles may exceed their limits before the posture can be recovered while gripping the same foothold. Therefore, it is beneficial to combine this approach with a control strategy that allows the robot to recover its balance by stepping in response to significant external forces even within a limited foothold~\cite{kojio2020footstep}.
Moreover, we plan to implement the proposed method in robots and verify its effectiveness through practical experiments.

%
%

\section*{Acknowledgments}
This work was supported by JST SPRING Grant Number JPMJSP2114 and JSPS KAKENHI Grant Number JP23K13281.

\end{document}